\documentclass{article}

\usepackage{arxiv}

\usepackage[utf8]{inputenc} 
\usepackage[T1]{fontenc}    
\usepackage{hyperref}       
\usepackage{url}            
\usepackage{booktabs}       
\usepackage{amsfonts}       
\usepackage{nicefrac}       
\usepackage{microtype}      
\usepackage{amsmath}
\usepackage{lipsum}
\usepackage{graphicx}
\graphicspath{ {./images/} }

\title{JoyStreamer-Flash: Real-time and Infinite Audio-Driven Avatar Generation with Autoregressive Diffusion}

\author{
\textbf{Chaochao Li} \quad \textbf{Ruikui Wang} \quad \textbf{Liangbo Zhou} \quad \textbf{Jinheng Feng} \quad \textbf{Huaishao Luo} \\
\textbf{Huan Zhang$^{\ddagger}$} \quad \textbf{Youzheng Wu} \quad \textbf{Xiaodong He} \\
\\
JD Technology \\
{\small $\ddagger$ Project Leader.}
}

\begin{document}
\maketitle

\begin{figure}[htbp]
\centering
\includegraphics[width=0.9\textwidth]{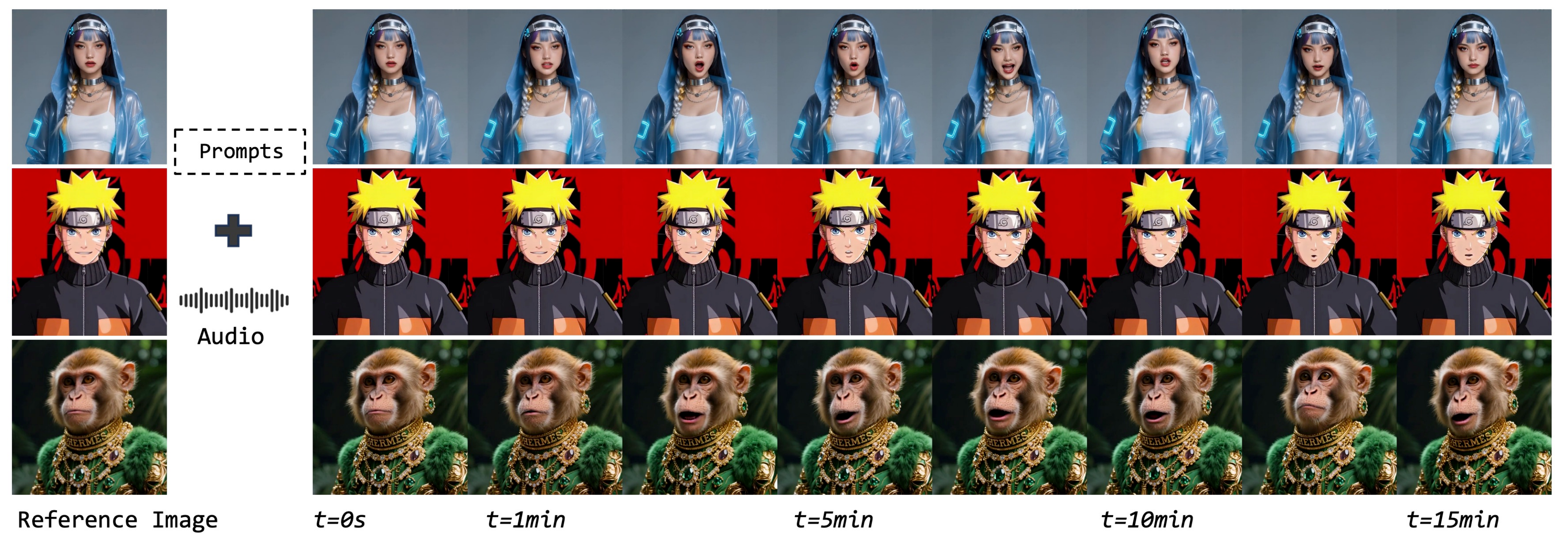}
\caption{JoyStreamer-Flash achieves real-time, streaming audio-driven avatar video generation at 16 FPS on a single GPU, while supporting infinite-length video synthesis with minimal error accumulation.}
\label{fig:abstract}
\end{figure}

\begin{abstract}
Existing DiT-based audio-driven avatar generation methods have achieved considerable progress, yet their broader application is constrained by limitations such as high computational overhead and the inability to synthesize long-duration videos. Autoregressive methods address this problem by applying block-wise autoregressive diffusion methods. However, these methods suffer from the problem of error accumulation and quality degradation. To address this, we propose \textbf{JoyStreamer-Flash}, an audio-driven autoregressive model capable of real-time inference and infinite-length video generation with the following contributions: (1) Progressive Step Bootstrapping (PSB), which allocates more denoising steps to initial frames to stabilize generation and reduce error accumulation; (2) Motion Condition Injection (MCI), enhancing temporal coherence by injecting noise-corrupted previous frames as motion condition; and (3) Unbounded RoPE via Cache-Resetting (URCR), enabling infinite-length generation through dynamic positional encoding. Our 1.3B-parameter causal model achieves 16 FPS on a single GPU and achieves competitive results in visual quality, temporal consistency, and lip synchronization.
\end{abstract}

\section{Introduction}
Talking avatars \cite{meng2025echomimicv3,cui2025hallo4,gan2025omniavatar,wang2025fantasytalking,tu2025stableavatar,gao2025wan}  synthesize identity-preserving, lip-synced, and motion-controllable videos based on an avatar representation, a driving audio signal, and textual prompts. While current mainstream DiT-based video generation methods \cite{yang2025infinitetalk,kong2412hunyuanvideo,wan2025wan} produce vivid and detailed outputs, their widespread adoption is significantly hindered by high computational costs, slow inference speeds, and limitations on short video length. In contrast, autoregressive video generation approaches \cite{yin2025slow,huang2025self,cui2025self,liu2025rolling} offer faster generation and support for long sequences. However, these methods often suffer from error accumulation inherent in the autoregressive process, leading to common artifacts such as inter-frame flickering and block-wise temporal inconsistency.

To address these challenges, we present JoyStreamer-Flash, a 1.3B-parameter, audio-driven model employing a 4-step denoising process within a causal architecture. Our model enables real-time, streaming generation with infinite length, audio-synchronized talking videos. It is designed to maintain high visual clarity and fidelity, ensure consistent identity preservation, and achieve robust temporal coherence throughout extended sequences.

The key contributions of our work are as follows:

\begin{itemize}
    \item \textbf{Progressive Step Bootstrapping (PSB).} We allocate more denoising steps to the critical initial frames at the beginning of the sequence to guide and stabilize the entire subsequent generation process. During training, the generator employs a dual-branch architecture, where one branch is randomly selected at each step to perform roll-out generation. The Auxiliary-Branch facilitates enhanced training stability for the Main-Branch. The denoising procedure is decomposed into Main-step Denoising and Sub-step Denoising, with the latter being stochastically optional. During inference, the first few blocks incorporate additional sub-step denoising, which enhances the synthesis quality of the initial frames, effectively mitigates autoregressive error accumulation, enriches fine-grained visual details, and maintains relatively high inference efficiency.

    \item \textbf{Motion Condition Injection (MCI).} Video sequences generated in an autoregressive block-by-block manner \cite{yin2025slow,huang2025self} often suffer from subtle yet perceptible temporal flickering at the pixel level. To address this inter-block temporal inconsistency, we introduce a novel conditioning mechanism: before each denoising window, we prepend a noise-corrupted version of the clean frame predicted at the previous timestep. This condition frame preserves and propagates high-frequency details from the preceding block, thereby significantly enhancing temporal smoothness across block boundaries and effectively alleviating or even eliminating temporal flicker between blocks.

    \item \textbf{Unbounded RoPE via Cache-Resetting (URCR). } We designate the first block as sink frames \cite{xiao2024efficient} to provide global reference information. While prior works fix the RoPE \cite{su2024roformer} indices of sink frames to start from zero \cite{huang2025self,low2025talkingmachines}, we instead apply relative positional encoding to the sink frames so that they remain temporally aligned with the subsequent local context. Furthermore, we store key (K) states in the KV cache before RoPE application and dynamically apply relative positional encoding at retrieval stage. This design enables flexible and consistent positional modeling over arbitrary lengths, empowering our model to generate videos of unbounded duration.
\end{itemize}

\section{Related Work}
\label{sec:headings}

\subsection{Audio-Driven Avatar Video Generation}
Audio-Driven Avatar Video Generation aims to synthesize realistic and expressive talking-head videos conditioned on an input speech signal and a reference avatar image. Recent advances \cite{meng2025echomimicv3,cui2025hallo4,gan2025omniavatar,wang2025fantasytalking,tu2025stableavatar,gao2025wan} have leveraged diffusion models for end-to-end audio-driven synthesis. These methods incorporate speech directly as a conditioning signal within a Diffusion Transformer (DiT) framework \cite{peebles2023scalable}, employing a unified attention mechanism to achieve joint alignment of audio content, facial expressions, and head movements. As a result, they achieve highly realistic and temporally coherent video generation without relying on intermediate representations, while also demonstrating strong performance in lip-sync accuracy and fine-grained expressive detail.

Despite these advances, existing DiT-based approaches remain constrained by practical limitations—notably high computational overhead and an inability to generate extended video sequences—which impedes their scalability and real-world applicability. In this work, we aim to develop a video generation model that achieves high visual fidelity, robust motion consistency, effective textual controllability, and sufficiently fast inference to enable real-time interactive applications.


\subsection{Autoregressive Long Video Generation}
To enable long-duration video synthesis, recent works have transitioned from bidirectional generation paradigms to autoregressive frameworks, which inherently support sequential unfolding over extended temporal horizons. Conventional autoregressive models are typically trained with a next-token prediction objective and generate spatiotemporal tokens in a strictly sequential manner during inference \cite{chen2024diffusion}.

A more recent line of research \cite{yin2025slow,huang2025self,cui2025self,liu2025rolling} integrates autoregressive modeling with denoising diffusion processes. In such hybrid approaches, an outer autoregressive loop iteratively generates video blocks, while an inner denoising loop progressively refines each frame within a block. For instance, SkyReels-V2 [29] combines diffusion forcing with reinforcement learning and employs a non-decreasing noise schedule to theoretically support infinite-length video synthesis, albeit at the cost of substantial computational overhead during training.

CausVid \cite{yin2025slow} leverages DMD distillation \cite{yin2024one} together with KV-cache mechanisms to accelerate generation and enable streaming inference, yet suffers from over-exposure artifacts due to error accumulation over long sequences. Self-Forcing \cite{huang2025self} mitigates exposure bias by conditioning on previously generated frames during training, thereby reducing the train-inference discrepancy. Its extension, Self-Forcing++ \cite{cui2025self}, further adapts the DMD framework to generate videos beyond the original temporal capacity of the base model. LongLive \cite{yang2025longlive} introduces KV re-caching and short-window attention to facilitate streaming and support long-context fine-tuning. Rolling-Forcing \cite{liu2025rolling} proposes a rolling-window joint denoising and attention sink mechanism \cite{xiao2024efficient} to suppress error propagation and enhance inter-block coherence.

Despite these advances, most existing methods still struggle with exposure bias and compounding errors when synthesizing long videos, often manifesting as visual artifacts or temporal inconsistencies. In contrast, our approach enables real-time, streaming-capable generation of infinite-length videos while simultaneously minimizing error accumulation and effectively suppressing inter-block temporal flickering.

\paragraph{Remark.}A concurrent work LiveAvatar \cite{huang2025live} presents a high-quality audio-driven avatar generation system based on a 14B-parameter causal model, achieving compelling visual fidelity and fast inference through multi-GPU parallelization and streaming-aware architectural design. LiveAvatar improves overall FPS via multi-GPU parallelism but suffers from high initial frame latency which limits rapid response to user inputs and constrains real-time interactivity. In contrast, our 1.3B causal model—with its lightweight architecture and joint multi‑block denoising design—avoids KV‑recaching overhead and thus supports interactive responsiveness.

\section{Methods}
We first provide background on autoregressive video diffusion models in Section 3.1. In Section 3.2, we adapt the Wan2.1 I2V architecture \cite{wan2025wan} to train an audio-conditioned bidirectional DiT for avatar video generation. We then introduce three key technical contributions: Progressive Step Bootstrapping (PSB) in Section 3.3, Motion Condition Injection (MCI) in Section 3.4, and Unbounded RoPE via Cache-Resetting (URCR) in Section 3.5.

\begin{figure}[h]
\centering
\includegraphics[width=0.9\textwidth]{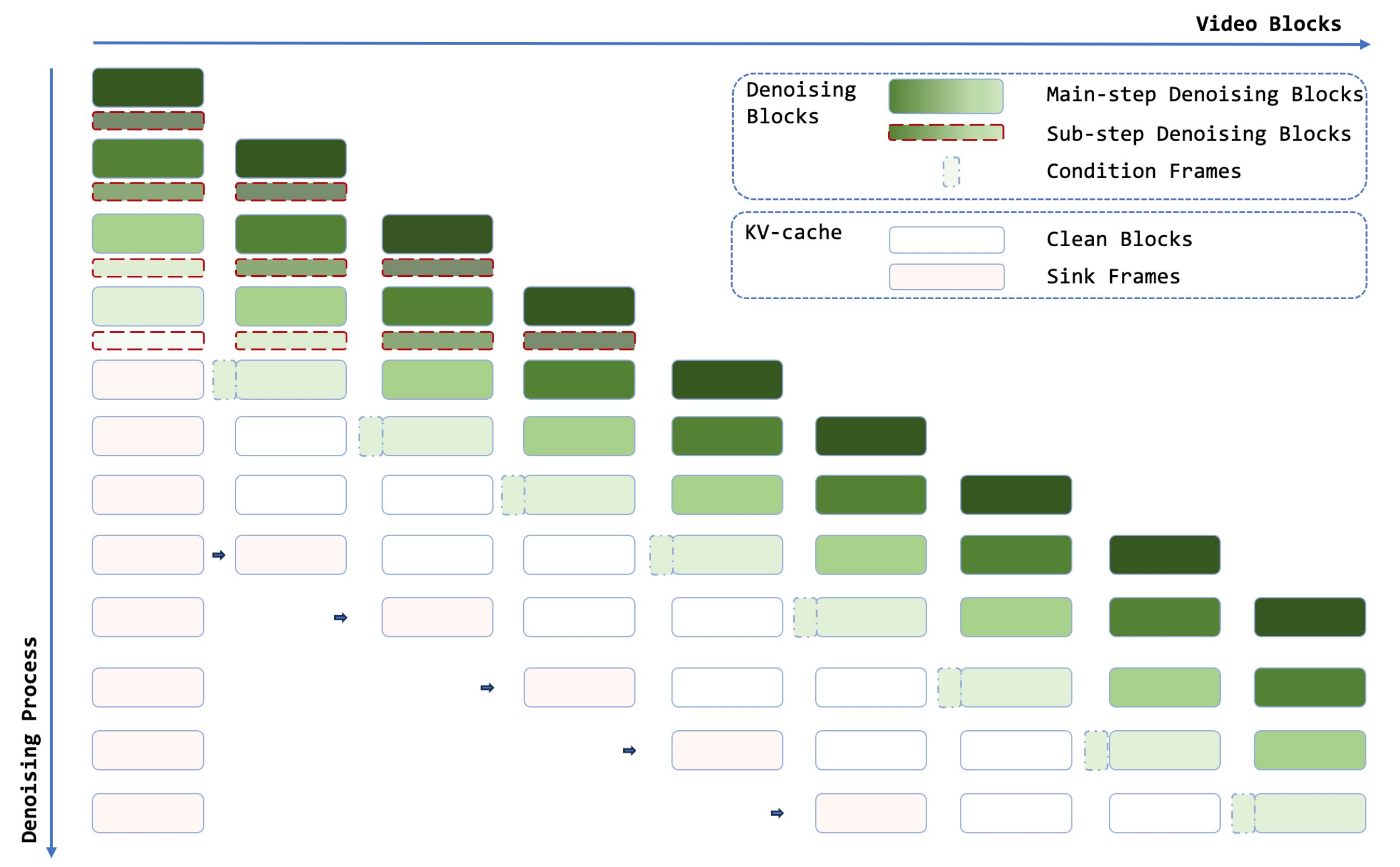}
\caption{\textbf{Overview of JoyStreamer-Flash.} \textit{Sub-step Denosing Blocks} implement \textbf{Progressive Step Bootstrapping (PSB)}, \textit{Condition Frames} embody \textbf{Motion Condition Injection (MCI)}, while \textit{Sink Frames connected via preceding arrows} demonstrate the mechanism of \textbf{Unbounded RoPE via Cache-Resetting (URCR)}.}
\label{fig:overview}
\end{figure}

\subsection{Preliminaries}
\paragraph{Autoregressive Video Diffusion Models.} An autoregressive video diffusion model generates each video block sequentially through a progressive denoising process, conditioned on previously generated frames. Given an input video sequence comprising \(N\) frames denoted as \(x^{1:N} = (x^1, x^2, \dots, x^N)\), the joint probability distribution can be factorized autoregressively via the chain rule:
\begin{equation}
    p(x^{1:N}) = \prod_{i=1}^{N} p\left(x^i \mid x^{1:i-1}\right)
    \label{eq:chain_rule}
\end{equation}

Here, each conditional term \(p(x^i \mid x^{1:i-1})\) is modeled using a diffusion process, where the generation of the \(i\)-th frame depends explicitly on all preceding frames \(x^{1:i-1} = (x^1, x^2, \dots, x^{i-1})\).

\paragraph{Distribution Matching Distillation (DMD).} DMD \cite{yin2024one,yin2024improved} reduces the computational cost of multi-step sampling in diffusion models by distilling knowledge into efficient few-step generators. Unlike conventional distillation approaches that replicate specific sampling trajectories, DMD aligns probability distributions, offering greater architectural freedom in designing the student network.

The core objective of DMD is to minimize the reverse KL divergence between the distribution produced by a multi-step teacher model and that of a parameter-efficient student generator. This objective is formally defined as:

\begin{equation}
    \mathcal{L}_{\text{DMD}} = \mathbb{E}_{t \sim \mathcal{T}} \left[ \mathrm{KL}\left( p_{\text{gen},t} \,\middle\|\, p_{\text{data},t} \right) \right]
    \label{eq:dmd_objective}
\end{equation}

The gradient of this loss function can be approximated in closed form.A practical and tractable gradient estimate is then given by:

\begin{equation}
    \nabla_{\phi} \mathcal{L}_{\text{DMD}} \approx -\mathbb{E}_t \left[ \int_{\epsilon} \Bigl( s_{\text{data}}\bigl(\Psi(G_\phi(\epsilon), t), t\bigr) - s_{\text{gen}}\bigl(\Psi(G_\phi(\epsilon), t), t\bigr) \Bigr) \cdot \frac{\partial G_\phi(\epsilon)}{\partial \phi} \, d\epsilon \right]
    \label{eq:dmd_gradient_approx}
\end{equation}

Here, $G_\phi$ represents the student generator with parameters $\phi$, $\Psi(\cdot, t)$ denotes the forward diffusion process at noise level $t$, and $s_{\text{data}}$ and $s_{\text{gen}}$ correspond to the score functions derived from the data distribution and the generator's output distribution, respectively.

CausVid \cite{yin2025slow} employs a diffusion forcing scheme \cite{chen2024diffusion} combined with DMD \cite{yin2024one, yin2024improved} to train a few-step causal diffusion model. The training utilizes a DMD loss that minimizes the reverse KL divergence between the data distribution $P_{\text{data}}$ and the generator’s output distribution $P_{\text{gen}}$ across randomly sampled timesteps $t$. The gradient of the reverse KL divergence can be expressed in closed form as the difference between two score functions.

Each denoising step is conditioned on both the current denoising window (e.g., 1 block in Self-Forcing \cite{huang2025self} or $T$ blocks in Rolling-Forcing \cite{liu2025rolling} and the KV-cache from previously generated frames.

Methods in this category typically maintain a KV-cache of fixed length $L$, which stores key-value pairs from both the most recent and earliest generated frames. Advancements in these approaches often focus on the design of the denoising window, the update strategy of the KV-cache, the application strategy of RoPE, and diverse training methodologies.

\subsection{Audio-Driven Bidirectional Model}
\begin{figure}[htbp]
\centering
\includegraphics[width=0.9\textwidth]{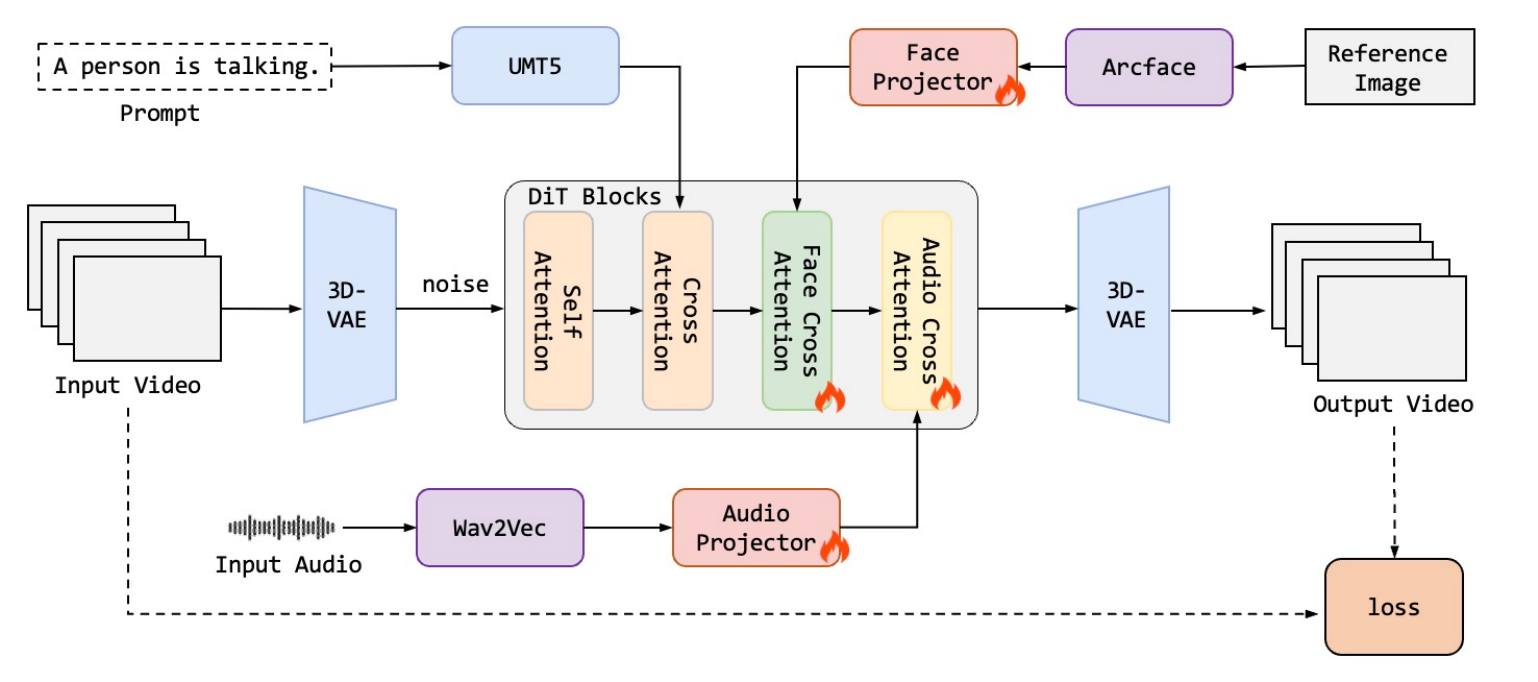}
\caption{The Training Framework of Audio-Driven Bidirectional DiT Model.}
\label{fig:bi-model}
\end{figure}
Based on the Wan2.1 I2V model \cite{wan2025wan}, we train an audio-driven bidirectional DiT model. The training framework is illustrated in Figure \ref{fig:bi-model}. First, audio features are extracted using Wav2Vec \cite{schneider2019wav2vec} and injected into DiT blocks through an audio projector and cross-attention layers, ensuring accurate lip synchronization in the generated videos. Second, facial identity features are obtained from the reference image via an ArcFace \cite{deng2019arcface} extractor and projected into the DiT blocks through a face projector and cross-attention mechanism to maintain identity consistency \cite{wang2025fantasytalking}. Third, we train both 14B and 1.3B audio-driven models from the corresponding Wan2.1 I2V backbones. In the subsequent DMD training stage, the 14B model serves as the real-score reference, while the 1.3B model acts as both the fake-score model and the initialized generator.

While bidirectional DiT architectures offer strong modeling capacity, they are limited by high computational overhead and inherently lack the temporal causality required for efficient streaming and long-form generation. Autoregressive approaches such as CausVid \cite{yin2025slow} and Self-Forcing \cite{huang2025self} address this by applying block-wise autoregressive diffusion, but still suffer from error accumulation and quality degradation over extended sequences. To overcome these limitations, we introduce a causal DiT framework with three key technical contributions: Progressive Step Bootstrapping (PSB), Motion Condition Injection (MCI) and Unbounded RoPE via Cache-Resetting (URCR).

\subsection{Progressive Step Bootstrapping (PSB)}
\begin{figure}[htbp]
\centering
\includegraphics[width=0.9\textwidth]{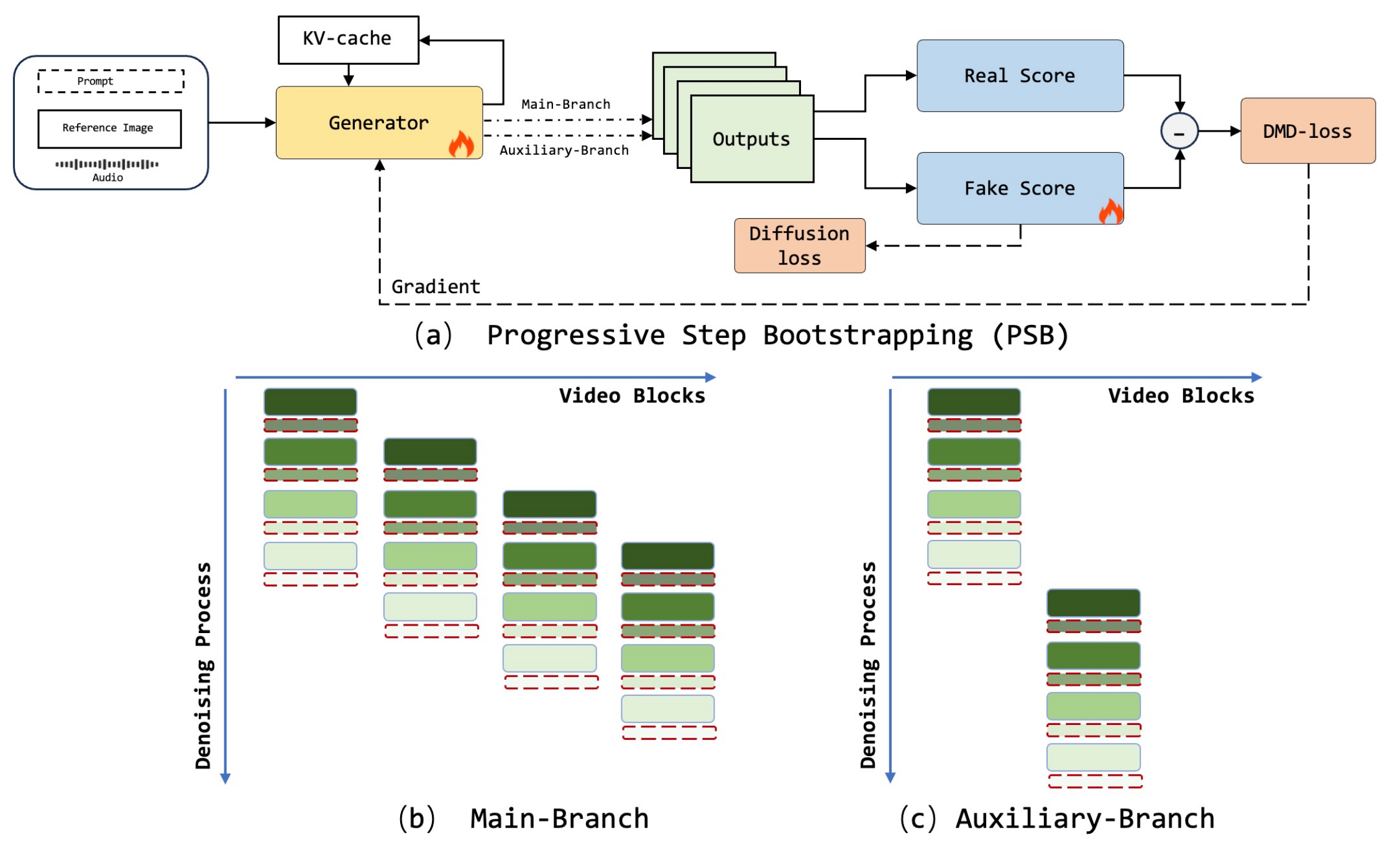}
\caption{(a) \textbf{Progressive Step Bootstrapping (PSB)} allocates additional denoising steps to critical initial frames to guide and stabilize subsequent generation. (b) \textbf{Main-Branch}  performs multi-block joint denoising with progressively scaled noise levels under the guidance of multi-step initialization from prior frames. (c) \textbf{Auxiliary-Branch} introduces intermediate sub-step denoising to provide denser supervision signals, thereby stabilizing training and enhancing the quality of preceding frames.}
\label{fig:method1}
\end{figure}
Prior work \cite{yang2025towards} notes that the initial frames in autoregressive DiT-based video generation are critical, as they serve as the conditioning context for all subsequent frames. Due to the lack of clean reference frames, errors introduced in early stages tend to propagate through the autoregressive process. Therefore, improving the quality of these initial frames is essential for overall generation stability. Our model operates with a denoising window of 4 blocks and a baseline of 4 denoising steps. To enhance the initial frames, we allocate additional denoising steps to the first 4 blocks during inference, thereby improving their stability, reducing artifacts, and mitigating error accumulation.

During training, we follow Rolling-Forcing \cite{liu2025rolling} by incorporating both joint denoising and block-wise denoising branches to ensure training stability. Distinctively, our approach integrates additional denoising steps into the training process. Specifically, the joint denoising branch performs main-branch denoising (as illustrated in Figure \ref{fig:method1}), progressively denoising multiple blocks under guided noise levels with enhanced early-frame conditioning. The block-wise branch introduces an auxiliary-branch denoising procedure, where additional sub-step denoising provides intermediate supervisory signals to stabilize training and improve initial frame quality.

We define main-step denoising with timestep indices [1000, 750, 500, 250] and sub-step denoising with indices [875, 625, 375, 125], corresponding to the midpoints between main steps. While main steps are always applied, sub-steps are stochastically selected during training via a gradient truncation strategy to ensure they receive sufficient supervision.

During inference, the first block is denoised 8 timesteps, the second 7 timesteps, the third 6 timesteps, and the fourth 5 timesteps. All subsequent blocks revert to the baseline 4-step denoising. After each denoising step, a clean block is generated and a new fully noisy block is appended. This progressive allocation of denoising steps results in videos with significantly improved stability, detail, and reduced artifact propagation.

\subsection{Motion Condition Injection (MCI)}
During the inference stage, videos generated using Progressive Step Bootstrapping (PSB) significantly mitigate autoregressive error accumulation, enabling stable rollouts at minute-scale durations while preserving motion dynamics and identity consistency. Nevertheless, subtle yet perceptible pixel-level temporal flickering persists at the boundaries between consecutive video blocks.

To alleviate such flickering between consecutive blocks, we innovatively introduce a conditioning frame prepended to the denoising window, which is derived from the clean frame predicted at the previous timestep and subsequently perturbed with noise. We term this strategy Motion Condition Injection (MCI). Specifically, we take the last latent of the preceding clean block, apply noise corresponding to the denoising step of the first block in the current window, and use the noised version as the conditioning frame. During training, noise is added to the conditioning frame with a probability of p = 0.7. Experiments demonstrate that this approach enhances temporal smoothness across block boundaries, effectively reduces inter‑block flickering and strengthens overall temporal coherence.

\subsection{Unbounded RoPE via Cache-Resetting (URCR)}
\begin{figure}[htbp]
\centering
\includegraphics[width=0.9\textwidth]{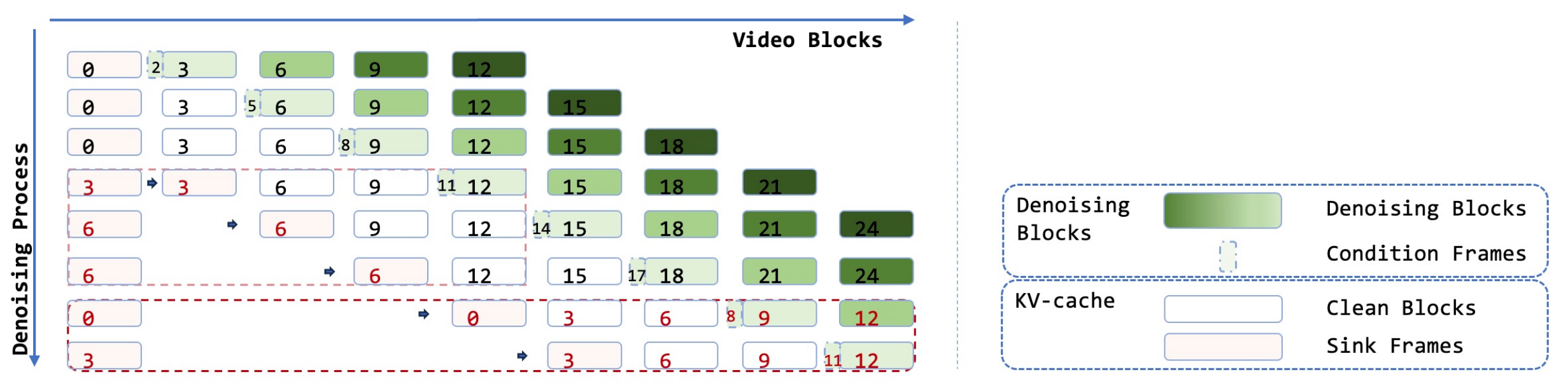}
\caption{\textbf{Unbounded RoPE via Cache-Resetting (URCR)} overcomes fixed-length constraints through cache resetting, with the numerical values inside each rectangular block indicating the RoPE index corresponding to its starting position.}
\label{fig:method3}
\end{figure}
One advantage of the AR-DiT framework is its ability to iteratively generate new video blocks based on previously synthesized content, theoretically enabling the creation of infinite-length videos. However, in practice, long-term generation is constrained not only by accumulated error but also by the positional limitations of Rotary Position Embeddings (RoPE) \cite{su2024roformer}. Typically, RoPE is configured with a fixed maximum range during training. For instance, the Wan video generation models \cite{wan2025wan} restricts the maximum sequence length to 1024 latents \cite{yesiltepe2025infinity}.

To overcome this constraint, we introduce several modifications to the standard RoPE mechanism. First, we designate the initial block as Sink Frames \cite{xiao2024efficient}, which serve as a global reference to maintain overall consistency. While some prior works \cite{low2025talkingmachines} fix the RoPE indices of sink frames to start from 0, we apply relative positional encoding to them, aligning their temporal positioning with subsequent local contexts, as illustrated in the central region of Figure \ref{fig:method3}.

Furthermore, to support video generation of arbitrary duration, we implement Unbounded RoPE via Cache-Resetting (URCR). Unlike conventional approaches that cache K-states after applying RoPE, URCR stores K-states before RoPE transformation in the KV-cache. During attention computation, when historical KV states are retrieved, relative positional encoding is dynamically applied to the cached K-values. Since the cached K-states remain position-agnostic, RoPE indices can be flexibly assigned during inference, thereby circumventing fixed-length limitations. Specifically, we define a maximum threshold for the RoPE index N. When the position of the denoising window exceeds N (N = 27 in Figure \ref{fig:method3}), the RoPE indices of the sink frames are reset to 0, and subsequent frames are indexed accordingly, as shown in the lower section of Figure 6. By removing the 1024-latent length constraint, our method theoretically enables the generation of indefinitely long videos.

\section{Experiment}
\subsection{Implementation Details}
\paragraph{Model.} We train audio-driven models of 14B and 1.3B parameters respectively, based on the Wan2.1 14B and 1.3B I2V architectures \cite{wan2025wan}. The 14B audio-driven model serves as the Real Score model in the subsequent DMD training stage, while the 1.3B audio-driven model functions as both the Fake Score model and the initialized Generator. Following the approach of CausVid \cite{yin2025slow}, we first initialize the 1.3B base model using causal attention masking on 10k ODE pairs sampled from the 14B audio-driven base model.

During DMD training, the model is trained on 64 GPUs for 10,000 steps, with a batch size of 1 per GPU. The joint denoising branch operates with a window length of 4 blocks, where each block contains 3 latent frames, resulting in a maximum sequence length of 27 latent frames during training. We employ the AdamW optimizer \cite{adam2014method} ($\beta_1 = 0.9$, $\beta_2 = 0.999$) for both the generator (learning rate $2 \times 10^{-6}$) and the fake score model (learning rate $4.0 \times 10^{-7}$). The fake score model is updated once every 5 generator updates.

\paragraph{Metrics.} To evaluate our model, we employ the Q-align \cite{wu2023q} framework to assess perceptual image quality (Q\_score) and perceptual face quality (Q\_face) of the synthesized videos. For audio-visual synchronization evaluation, we employ Sync-C and Sync-D metrics \cite{chung2016out} to assess the temporal alignment between generated lip movements and the input audio signal. To evaluate identity consistency in the generated videos, we adopt DINOv2 \cite{oquab2023dinov2} as our metric, which we refer to as Identity Consistency (IDC). On our custom out-of-distribution (OOD) test set, which lacks ground-truth videos, we report only the Q\_score, Q\_face, Sync-C, Sync-D, and IDC scores.

\paragraph{Evaluation Data.} We construct a custom evaluation dataset comprising 100 diverse avatars, including real human subjects, animated characters, and 3D models. Each avatar is paired with speech audio drawn from multiple domains, including conversational dialogue, public speaking, singing, traditional opera, and TTS-generated utterances.

\subsection{Comparison}
We evaluate JoyStreamer-Flash and its bidirectional teacher model against state-of-the-art audio-driven video generation methods. Quantitative results across key metrics are presented in Table \ref{tab:comparison_detailed}.

\begin{table}[htbp]
\centering
\caption{Comparison of audio-driven video generation models across open-source, closed-source, and causal categories. FPS is measured under single‑GPU inference. As Omnihuman‑1 and Heygen are closed‑source, their model scale and actual single‑GPU inference speed cannot be precisely determined. To ensure a fair comparison, we focus on evaluating models that share the same causal DiT architecture. Accordingly, the bolded metrics in the table highlight comparisons specifically among causal-based audio-driven models.}
\label{tab:comparison_detailed}
\begin{tabular}{l l c c c c c c}
\toprule

\multicolumn{2}{c}{Model} & \multicolumn{6}{c}{Metrics} \\
\cmidrule(lr){3-8}
& & Sync-C$\uparrow$ & Sync-D$\downarrow$ & IDC$\uparrow$ & Qscore$\uparrow$ & Q\_face$\uparrow$ & FPS$\uparrow$ \\
\midrule

\multicolumn{8}{l}{\small\textit{Open-Source Audio-driven Bidirectional DiT models}} \\
StableAvatar~\cite{tu2025stableavatar} & 1.3B & 2.47 & 10.97 & 0.72 & 0.91 & 0.53 & $<1$ \\
EchoMimicV3~\cite{meng2025echomimicv3} & 1.3B & 2.49 & 10.78 & 0.72 & 0.90 & 0.51 & $<1$ \\
OmniAvatar~\cite{gan2025omniavatar} & 1.3B & 3.85 & 9.54 & 0.72 & 0.88 & 0.50 & $<1$ \\
WanS2V~\cite{gao2025wan} & 14B & 4.05 & 9.38 & 0.67 & 0.88 & 0.47 & $<1$ \\
Our teacher model & 14B & 5.56 & 8.19 & 0.78 & 0.91 & 0.63 & $<1$ \\
\midrule

\multicolumn{8}{l}{\small\textit{Closed-Source Audio-driven models}} \\  
Omnihuman-1~\cite{lin2025omnihuman} & -- & 5.62 & 8.80 & 0.82 & 0.90 & 0.65 & -- \\
Heygen~\cite{heygen2024} & -- & 4.86 & 9.09 & 0.79 & 0.92 & 0.66 & -- \\
\midrule

\multicolumn{8}{l}{\small\textit{Audio-driven Causal DiT models}} \\
LiveAvatar~\cite{huang2025live} & 14B & 3.89 & 9.45 & \textbf{0.78} & \textbf{0.85} & 0.52 & 4.3 \\
Ours & 1.3B & \textbf{5.25} & \textbf{8.30} & 0.72 & 0.84 & \textbf{0.53} & \textbf{16.2} \\
\bottomrule
\end{tabular}
\end{table}

As shown in Table \ref{tab:comparison_detailed}, our bidirectional teacher model achieves superior performance across all evaluated metrics—including video quality, lip‑sync accuracy, and identity preservation—compared to existing open‑source methods. When distilled into a causal variant via Diffusion Model Distillation (DMD), the resulting JoyStreamer-Flash model shows a moderate performance decline, which is attributable to its reduced parameter scale and the constraints of its causal architecture. Nonetheless, JoyStreamer-Flash maintains competitive performance across all metrics relative to bidirectional models of comparable scale. Notably, it significantly outperforms the 14B‑parameter causal model LiveAvatar \cite{huang2025live} in terms of Sync‑C and Sync‑D scores, while delivering comparable results on IDC, Q\_score, and Q\_face metrics. Importantly, the causal design yields a substantial improvement in inference speed: the DiT component reaches 16.2 FPS on a single GPU. With multi‑GPU parallelization and streaming optimizations, the full pipeline exceeds 30 FPS, enabling real‑time display and interactive applications.

\subsection{Ablation Study}
\begin{table}[h]
\caption{Ablation Study on Progressive Step Bootstrapping (PSB)}
\label{tab:ablation_PSB}
\centering
\begin{tabular}{lcccccc}
\toprule
\multicolumn{1}{c}{Methods} & \multicolumn{5}{c}{Metrics} \\
\cmidrule{2-6}
& Sync-C$\uparrow$ & Sync-D$\downarrow$ & IDC$\uparrow$ & Qscore$\uparrow$ & Q\_face$\uparrow$ \\
\midrule
w/o main-branch & 4.09 & 9.32 & \textbf{0.76} & 0.82 & 0.52 \\
w/o auxiliary-branch & 5.24 & 8.35 & 0.70 & 0.80 & 0.51 \\
w/o sub-step & 5.24 & 8.32 & 0.71 & 0.82 & 0.51 \\
Our full & \textbf{5.25} & \textbf{8.30} & 0.72 & \textbf{0.84} & \textbf{0.53} \\
\bottomrule
\end{tabular}
\end{table}
\begin{figure}[htbp]
\centering
\includegraphics[width=0.9\textwidth]{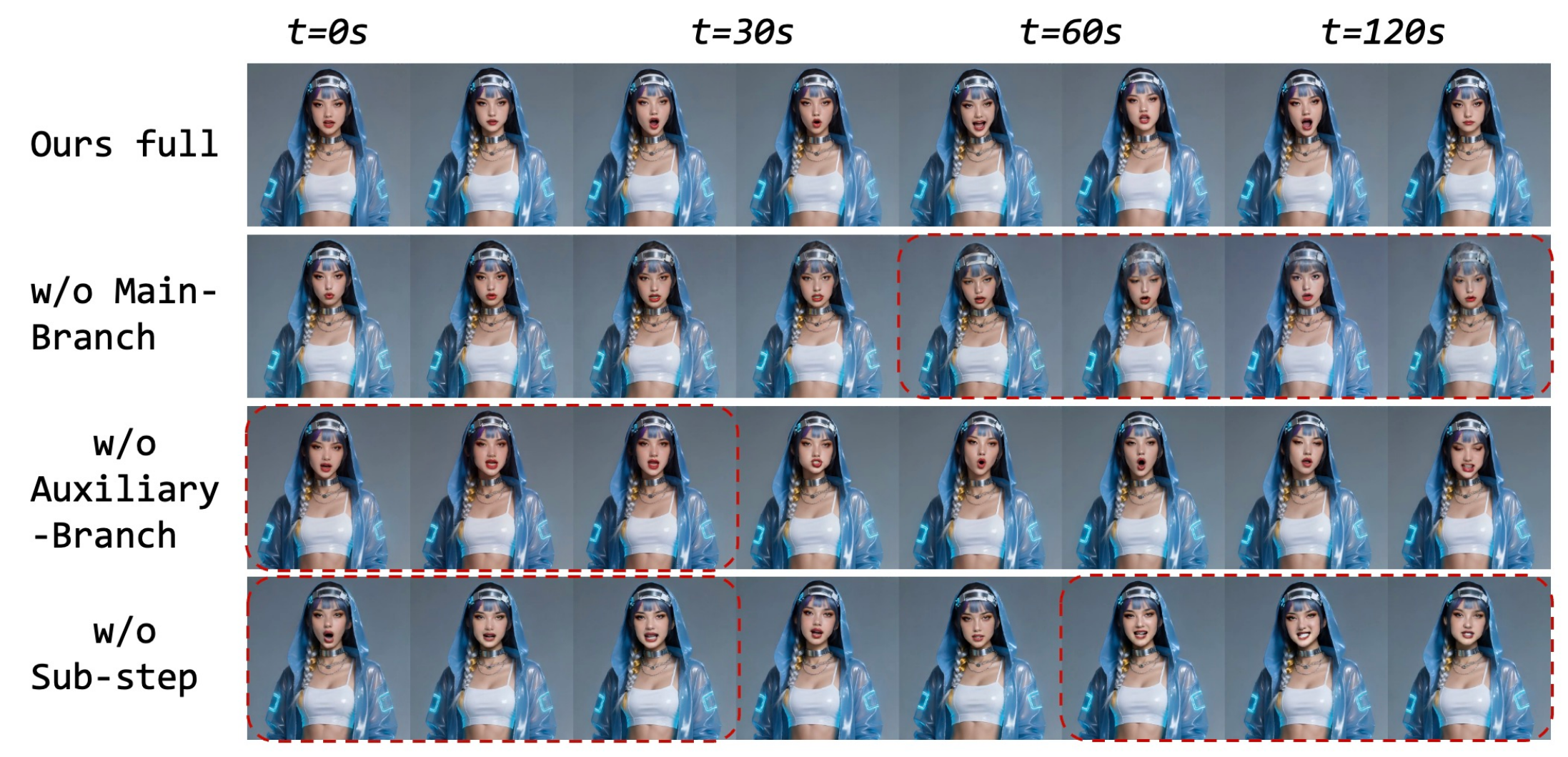}
\caption{Ablation Study on Progressive Step Bootstrapping (PSB).}
\label{fig:ablation1}
\end{figure}

\paragraph{Progressive Step Bootstrapping (PSB).} To evaluate the contribution of PSB, we conduct ablation studies by training three model variants: (1) a model without BPD training, (2) a model without AMSS training, and (3) a model without sub-step training. As illustrated in Figure \ref{fig:ablation1}, the model without BPD training exhibits significant quality degradation, blurriness, and accumulated artifacts in the latter half of long-video inference. The model without AMSS training demonstrates noticeable color distortion and reduced temporal coherence. Meanwhile, the model without sub-step training shows inferior quality in the initial frames and visible artifacts in subsequent segments of the generated video.

\begin{table}[h]
\centering
\caption{Ablation Study on Motion Condition Injection (MCI)}
\label{tab:ablation_MCI}
\begin{tabular}{lccccc}
\toprule
\multicolumn{1}{c}{Methods} & \multicolumn{5}{c}{Metrics} \\
\cmidrule{2-6}
& Sync-C$\uparrow$ & Sync-D$\downarrow$ & IDC$\uparrow$ & Qscore$\uparrow$ & Q\_face$\uparrow$ \\
\midrule
w/o MCI training & 5.19 & 8.33 & 0.71 & 0.83 & 0.52 \\
w/o add-noise & 5.19 & 8.33 & \textbf{0.72} & 0.82 & 0.52 \\
Our full & \textbf{5.25} & \textbf{8.30} & \textbf{0.72} & \textbf{0.84} & \textbf{0.53} \\
\bottomrule
\end{tabular}
\end{table}
\begin{figure}[htbp]
\centering
\includegraphics[width=0.9\textwidth]{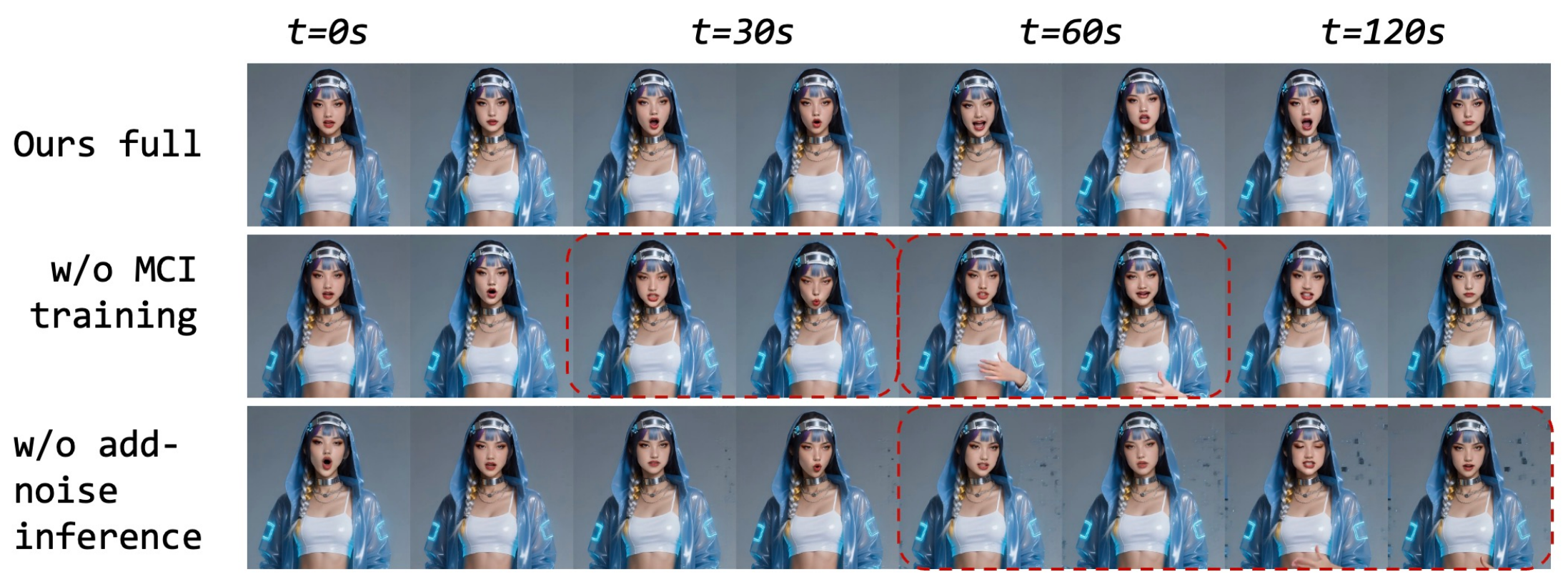}
\caption{Ablation Study on Motion Condition Injection (MCI).}
\label{fig:ablation2}
\end{figure}

\paragraph{Motion Condition Injection (MCI).} We compare models trained with and without MCI, as well as inference using condition frames without added noise (w/o add‑noise inference). As shown in Figure \ref{fig:ablation2}, videos generated without MCI during training exhibit noticeable modulation and flickering at block transitions. Inference without noise injection leads to pronounced error accumulation, particularly severe artifacts in later segments of the sequence. By incorporating MCI in both training and inference, temporal smoothness across block boundaries is significantly enhanced, effectively eliminating inter‑block flickering and artifacts.

\begin{table}[h]
\centering
\caption{Ablation Study on Unbounded RoPE via Cache-Resetting (URCR)}
\label{tab:ablation_URCR}
\begin{tabular}{lccccc}
\toprule
\multicolumn{1}{c}{Methods} & \multicolumn{5}{c}{Metrics} \\
\cmidrule{2-6}
& Sync-C$\uparrow$ & Sync-D$\downarrow$ & IDC$\uparrow$ & Qscore$\uparrow$ & Q\_face$\uparrow$ \\
\midrule
w/o sink frames & 5.03 & 8.43 & 0.65 & 0.58 & 0.37 \\
w/o URC & 5.20 & 8.32 & 0.72 & 0.83 & 0.52 \\
N-thread=100 & 5.24 & 8.31 & \textbf{0.72} & \textbf{0.84} & \textbf{0.53} \\
Our full & \textbf{5.25} & \textbf{8.30} & \textbf{0.72} & \textbf{0.84} & \textbf{0.53} \\
\bottomrule
\end{tabular}
\end{table}

\begin{figure}[htbp]
\centering
\includegraphics[width=0.9\textwidth]{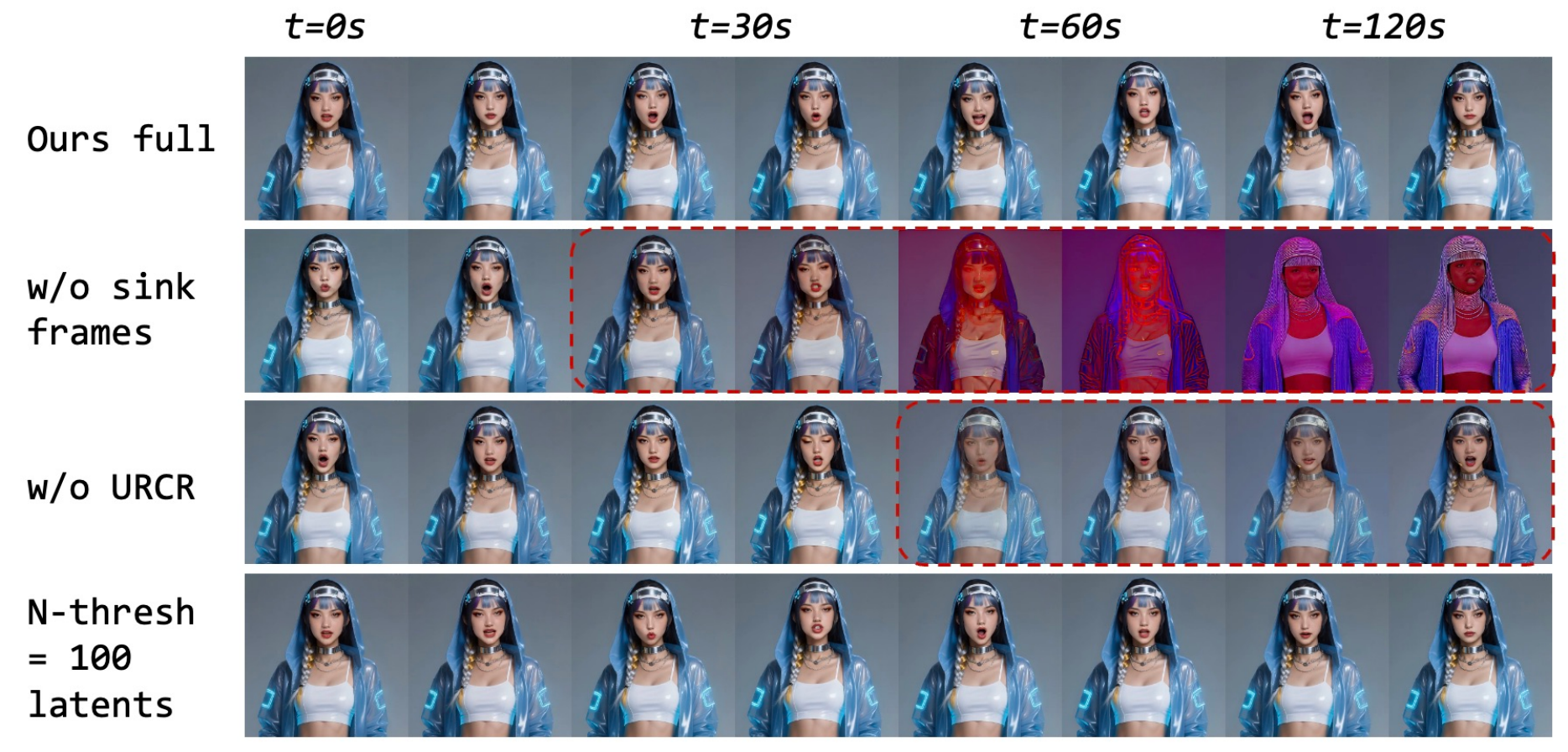}
\caption{Ablation Study on Unbounded RoPE via Cache-Resetting (URCR).}
\label{fig:ablation3}
\end{figure}
\paragraph{Unbounded RoPE via Cache-Resetting (URCR).} We compare the impact of omitting sink frames during inference, using sink frames without URCR, and applying the full URCR strategy. As shown in Figure \ref{fig:ablation3}, inference without sink frames leads to clear cumulative errors, while using sink frames without URCR results in significant quality degradation during long‑video generation. In contrast, the proposed URCR method enables theoretically unbounded‑length video synthesis.

We further investigate the influence of the maximum relative position encoding length threshold. Experiments show that whether the threshold is set to a small value (e.g., 100 tokens) or a large one (e.g., 1000 tokens), the inference results remain visually consistent. This indicates that any threshold within 1024 tokens effectively circumvents the RoPE length‑limitation issue, confirming the robustness of our approach.

\section{Limitation and Future Work}
Our generator model is derived from the Wan 1.3B I2V architecture. Given the inherent limitations in model capacity and inference capability, it exhibits suboptimal performance when processing complex scenes or fine-grained facial details. To address these constraints, future work will involve training the generator on a larger-scale model, with the goal of achieving stronger text alignment and enhanced detail fidelity.

\section{Conclusion}
We train an audio-conditioned bidirectional model based on the Wan2.1 I2V architecture and subsequently distill it into a causal generator using Denoising Matching Distillation (DMD). To mitigate error accumulation and temporal flickering inherent in asymmetric distillation, we introduce Progressive Step Bootstrapping (PSB), a training strategy that allocates additional denoising steps to the initial key frames of a sequence, thereby guiding and stabilizing the subsequent autoregressive generation process. Furthermore, we propose Motion Condition Injection (MCI), a training-and-inference-consistent technique that concatenates a noise-augmented clean frame from the previous timestep as a conditioning input, effectively preserving motion coherence while propagating high-frequency details to alleviate inter-block flickering artifacts. At inference stage, our Unbounded RoPE via Cache-Resetting (URCR) method stores K states without RoPE encoding in the KV cache and dynamically applies relative positional encoding during retrieval, enabling the generation of infinitely long videos while maintaining temporal consistency. Experimental results demonstrate that our approach achieves state-of-the-art performance in terms of temporal coherence, lip-sync accuracy, and inference speed.

\bibliographystyle{unsrt}  
\bibliography{main}  

@inproceedings{lin2025omnihuman,
  title={Omnihuman-1: Rethinking the scaling-up of one-stage conditioned human animation models},
  author={Lin, Gaojie and Jiang, Jianwen and Yang, Jiaqi and Zheng, Zerong and Liang, Chao and Zhang, Yuan and Liu, Jingtuo},
  booktitle={Proceedings of the IEEE/CVF International Conference on Computer Vision},
  pages={13847--13858},
  year={2025}
}

@article{meng2025echomimicv3,
  title={Echomimicv3: 1.3 b parameters are all you need for unified multi-modal and multi-task human animation},
  author={Meng, Rang and Wang, Yan and Wu, Weipeng and Zheng, Ruobing and Li, Yuming and Ma, Chenguang},
  journal={arXiv preprint arXiv:2507.03905},
  year={2025}
}

@article{cui2025hallo4,
  title={Hallo4: High-Fidelity Dynamic Portrait Animation via Direct Preference Optimization and Temporal Motion Modulation},
  author={Cui, Jiahao and Chen, Yan and Xu, Mingwang and Shang, Hanlin and Chen, Yuxuan and Zhan, Yun and Dong, Zilong and Yao, Yao and Wang, Jingdong and Zhu, Siyu},
  journal={arXiv preprint arXiv:2505.23525},
  year={2025}
}

@article{gan2025omniavatar,
  title={OmniAvatar: Efficient Audio-Driven Avatar Video Generation with Adaptive Body Animation},
  author={Gan, Qijun and Yang, Ruizi and Zhu, Jianke and Xue, Shaofei and Hoi, Steven},
  journal={arXiv preprint arXiv:2506.18866},
  year={2025}
}

@article{gao2025wan,
  title={Wan-s2v: Audio-driven cinematic video generation},
  author={Gao, Xin and Hu, Li and Hu, Siqi and Huang, Mingyang and Ji, Chaonan and Meng, Dechao and Qi, Jinwei and Qiao, Penchong and Shen, Zhen and Song, Yafei and others},
  journal={arXiv preprint arXiv:2508.18621},
  year={2025}
}

@inproceedings{wang2025fantasytalking,
  title={Fantasytalking: Realistic talking portrait generation via coherent motion synthesis},
  author={Wang, Mengchao and Wang, Qiang and Jiang, Fan and Fan, Yaqi and Zhang, Yunpeng and Qi, Yonggang and Zhao, Kun and Xu, Mu},
  booktitle={Proceedings of the 33rd ACM International Conference on Multimedia},
  pages={9891--9900},
  year={2025}
}

@article{tu2025stableavatar,
  title={Stableavatar: Infinite-length audio-driven avatar video generation},
  author={Tu, Shuyuan and Pan, Yueming and Huang, Yinming and Han, Xintong and Xing, Zhen and Dai, Qi and Luo, Chong and Wu, Zuxuan and Jiang, Yu-Gang},
  journal={arXiv preprint arXiv:2508.08248},
  year={2025}
}

@article{kong2412hunyuanvideo,
  title={Hunyuanvideo: A systematic framework for large video generative models, 2025},
  author={Kong, W and Tian, Q and Zhang, Z and Min, R and Dai, Z and Zhou, J and Xiong, J and Li, X and Wu, B and Zhang, J and others},
  journal={URL https://arxiv. org/abs/2412.03603}
}

@article{yang2025infinitetalk,
  title={Infinitetalk: Audio-driven video generation for sparse-frame video dubbing},
  author={Yang, Shaoshu and Kong, Zhe and Gao, Feng and Cheng, Meng and Liu, Xiangyu and Zhang, Yong and Kang, Zhuoliang and Luo, Wenhan and Cai, Xunliang and He, Ran and others},
  journal={arXiv preprint arXiv:2508.14033},
  year={2025}
}

@article{adam2014method,
  title={A method for stochastic optimization},
  author={Adam, Kingma DP Ba J and others},
  journal={arXiv preprint arXiv:1412.6980},
  volume={1412},
  number={6},
  year={2014}
}

@inproceedings{peebles2023scalable,
  title={Scalable diffusion models with transformers},
  author={Peebles, William and Xie, Saining},
  booktitle={Proceedings of the IEEE/CVF international conference on computer vision},
  pages={4195--4205},
  year={2023}
}

@article{wan2025wan,
  title={Wan: Open and advanced large-scale video generative models},
  author={Wan, Team and Wang, Ang and Ai, Baole and Wen, Bin and Mao, Chaojie and Xie, Chen-Wei and Chen, Di and Yu, Feiwu and Zhao, Haiming and Yang, Jianxiao and others},
  journal={arXiv preprint arXiv:2503.20314},
  year={2025}
}

@article{xiao2024efficient,
  title={Efficient streaming language models with attention sinks, 2024},
  author={Xiao, Guangxuan and Tian, Yuandong and Chen, Beidi and Han, Song and Lewis, Mike},
  journal={URL https://arxiv. org/abs/2309.17453},
  volume={1},
  year={2024}
}

@article{su2024roformer,
  title={Roformer: Enhanced transformer with rotary position embedding},
  author={Su, Jianlin and Ahmed, Murtadha and Lu, Yu and Pan, Shengfeng and Bo, Wen and Liu, Yunfeng},
  journal={Neurocomputing},
  volume={568},
  pages={127063},
  year={2024},
  publisher={Elsevier}
}

@article{chen2024diffusion,
  title={Diffusion forcing: Next-token prediction meets full-sequence diffusion},
  author={Chen, Boyuan and Mart{\'\i} Mons{\'o}, Diego and Du, Yilun and Simchowitz, Max and Tedrake, Russ and Sitzmann, Vincent},
  journal={Advances in Neural Information Processing Systems},
  volume={37},
  pages={24081--24125},
  year={2024}
}

@article{yin2024improved,
  title={Improved distribution matching distillation for fast image synthesis},
  author={Yin, Tianwei and Gharbi, Micha{\"e}l and Park, Taesung and Zhang, Richard and Shechtman, Eli and Durand, Fredo and Freeman, Bill},
  journal={Advances in neural information processing systems},
  volume={37},
  pages={47455--47487},
  year={2024}
}

@inproceedings{yin2024one,
  title={One-step diffusion with distribution matching distillation},
  author={Yin, Tianwei and Gharbi, Micha{\"e}l and Zhang, Richard and Shechtman, Eli and Durand, Fredo and Freeman, William T and Park, Taesung},
  booktitle={Proceedings of the IEEE/CVF conference on computer vision and pattern recognition},
  pages={6613--6623},
  year={2024}
}

@inproceedings{yin2025slow,
  title={From slow bidirectional to fast autoregressive video diffusion models},
  author={Yin, Tianwei and Zhang, Qiang and Zhang, Richard and Freeman, William T and Durand, Fredo and Shechtman, Eli and Huang, Xun},
  booktitle={Proceedings of the Computer Vision and Pattern Recognition Conference},
  pages={22963--22974},
  year={2025}
}

@article{huang2025self,
  title={Self Forcing: Bridging the Train-Test Gap in Autoregressive Video Diffusion},
  author={Huang, Xun and Li, Zhengqi and He, Guande and Zhou, Mingyuan and Shechtman, Eli},
  journal={arXiv preprint arXiv:2506.08009},
  year={2025}
}

@article{cui2025self,
  title={Self-Forcing++: Towards Minute-Scale High-Quality Video Generation},
  author={Cui, Justin and Wu, Jie and Li, Ming and Yang, Tao and Li, Xiaojie and Wang, Rui and Bai, Andrew and Ban, Yuanhao and Hsieh, Cho-Jui},
  journal={arXiv preprint arXiv:2510.02283},
  year={2025}
}

@article{liu2025rolling,
  title={Rolling forcing: Autoregressive long video diffusion in real time},
  author={Liu, Kunhao and Hu, Wenbo and Xu, Jiale and Shan, Ying and Lu, Shijian},
  journal={arXiv preprint arXiv:2509.25161},
  year={2025}
}

@article{yang2025longlive,
  title={Longlive: Real-time interactive long video generation},
  author={Yang, Shuai and Huang, Wei and Chu, Ruihang and Xiao, Yicheng and Zhao, Yuyang and Wang, Xianbang and Li, Muyang and Xie, Enze and Chen, Yingcong and Lu, Yao and others},
  journal={arXiv preprint arXiv:2509.22622},
  year={2025}
}

@article{low2025talkingmachines,
  title={TalkingMachines: Real-Time Audio-Driven FaceTime-Style Video via Autoregressive Diffusion Models},
  author={Low, Chetwin and Wang, Weimin},
  journal={arXiv preprint arXiv:2506.03099},
  year={2025}
}

@article{yesiltepe2025infinity,
  title={Infinity-RoPE: Action-Controllable Infinite Video Generation Emerges From Autoregressive Self-Rollout},
  author={Yesiltepe, Hidir and Meral, Tuna Han Salih and Akan, Adil Kaan and Oktay, Kaan and Yanardag, Pinar},
  journal={arXiv preprint arXiv:2511.20649},
  year={2025}
}

@online{heygen2024,
  author = {HeyGen},
  title = {AI Video Creation Platform | Heygen},
  year = {2024},
  url = {https://www.heygen.com/}
}

@article{huang2025live,
  title={Live Avatar: Streaming Real-time Audio-Driven Avatar Generation with Infinite Length},
  author={Huang, Yubo and Guo, Hailong and Wu, Fangtai and Zhang, Shifeng and Huang, Shijie and Gan, Qijun and Liu, Lin and Zhao, Sirui and Chen, Enhong and Liu, Jiaming and others},
  journal={arXiv preprint arXiv:2512.04677},
  year={2025}
}

@article{schneider2019wav2vec,
  title={wav2vec: Unsupervised pre-training for speech recognition},
  author={Schneider, Steffen and Baevski, Alexei and Collobert, Ronan and Auli, Michael},
  journal={arXiv preprint arXiv:1904.05862},
  year={2019}
}

@inproceedings{deng2019arcface,
  title={Arcface: Additive angular margin loss for deep face recognition},
  author={Deng, Jiankang and Guo, Jia and Xue, Niannan and Zafeiriou, Stefanos},
  booktitle={Proceedings of the IEEE/CVF conference on computer vision and pattern recognition},
  pages={4690--4699},
  year={2019}
}

@article{yang2025towards,
  title={Towards One-step Causal Video Generation via Adversarial Self-Distillation},
  author={Yang, Yongqi and Huang, Huayang and Peng, Xu and Hu, Xiaobin and Luo, Donghao and Zhang, Jiangning and Wang, Chengjie and Wu, Yu},
  journal={arXiv preprint arXiv:2511.01419},
  year={2025}
}

@article{wu2023q,
  title={Q-align: Teaching lmms for visual scoring via discrete text-defined levels},
  author={Wu, Haoning and Zhang, Zicheng and Zhang, Weixia and Chen, Chaofeng and Liao, Liang and Li, Chunyi and Gao, Yixuan and Wang, Annan and Zhang, Erli and Sun, Wenxiu and others},
  journal={arXiv preprint arXiv:2312.17090},
  year={2023}
}

@inproceedings{chung2016out,
  title={Out of time: automated lip sync in the wild},
  author={Chung, Joon Son and Zisserman, Andrew},
  booktitle={Asian conference on computer vision},
  pages={251--263},
  year={2016},
  organization={Springer}
}

@article{oquab2023dinov2,
  title={Dinov2: Learning robust visual features without supervision},
  author={Oquab, Maxime and Darcet, Timoth{\'e}e and Moutakanni, Th{\'e}o and Vo, Huy and Szafraniec, Marc and Khalidov, Vasil and Fernandez, Pierre and Haziza, Daniel and Massa, Francisco and El-Nouby, Alaaeldin and others},
  journal={arXiv preprint arXiv:2304.07193},
  year={2023}
}

\end{document}